\documentclass[nonacm,sigconf]{acmart}
\AtBeginDocument{%
  \providecommand\BibTeX{{%
    \normalfont B\kern-0.5em{\scshape i\kern-0.25em b}\kern-0.8em\TeX}}}

\usepackage{xspace}
\newcommand{\company}{CompanyX\xspace}

\usepackage{graphicx}
\usepackage{subcaption}

\usepackage{tabularx, booktabs}
\newcolumntype{Y}{>{\centering\arraybackslash}X}
\usepackage{multirow}
\usepackage{arydshln}

\usepackage[inline]{enumitem}

\usepackage{soul}

\setcopyright{none}
\begin{document}

\title{Leveraging GPT-4 Model on Root Cause Analysis \\through In-Context Learning}
\title[]{Automated Root Causing of Cloud Incidents using \\In-Context Learning with GPT-4}
\author{
Xuchao Zhang, Supriyo Ghosh, Chetan Bansal,
Rujia Wang, \\Minghua Ma, Yu Kang, Saravan Rajmohan \\
\vspace{0.1cm}
 Microsoft\\
 \vspace{0.1cm}
\{xuchaozhang, supriyoghosh, chetanb, rujiawang, minghuama, yu.kang, saravan.rajmohan\}@microsoft.com
}

\renewcommand{\shortauthors}{Zhang et al.}

\begin{abstract}
Root Cause Analysis (RCA) plays a pivotal role in the incident diagnosis process for cloud services, requiring on-call engineers to identify the primary issues and implement corrective actions to prevent future recurrences. Improving the incident RCA process is vital for minimizing service downtime, customer impact and manual toil. Recent advances in artificial intelligence have introduced state-of-the-art Large Language Models (LLMs) like GPT-4, which have proven effective in tackling various AIOps problems, ranging from code authoring to incident management. Nonetheless, the GPT-4 model's immense size presents challenges when trying to fine-tune it on user data because of the significant GPU resource demand and the necessity for continuous model fine-tuning with the emergence of new data. To address the high cost of fine-tuning LLM, we propose an in-context learning approach for automated root causing, which eliminates the need for fine-tuning. We conduct extensive study over 100,000 production incidents from \company{}, comparing several large language models using multiple metrics. The results reveal that our in-context learning approach outperforms the previous fine-tuned large language models such as GPT-3 by an average of 24.8\% across all metrics, with an impressive 49.7\% improvement over the zero-shot model. Moreover, human evaluation involving actual incident owners demonstrates its superiority over the fine-tuned model, achieving a 43.5\% improvement in correctness and an 8.7\% enhancement in readability. The impressive results demonstrate the viability of utilizing a vanilla GPT model for the RCA task, thereby avoiding the high computational and maintenance costs associated with a fine-tuned model.

\end{abstract}





\maketitle

\section{Introduction} \label{sec:introduction}

Over the last decade, the IT industry has transitioned away from the traditional practice of distributing software in shrink-wrapped packages. Instead, these companies are increasingly embracing cloud platforms as their favored approach for deploying applications and services. Within these extensive cloud services, incidents such as unplanned interruptions or performance degradation can have a significant negative impact on customer satisfaction, resulting in revenue loss and a decline in customer trust. At present, the process of diagnosing such incidents still heavily relies on manual investigation or the use of specialized service tools. Nevertheless, due to the escalating scale and complexity of modern cloud systems, relying solely on human capabilities is inadequate for effectively and efficiently handling incidents, leading to extended Time-to-Mitigate (TTM). 

Root cause analysis, as a pivotal task during incident management lifecycle \cite{chen2020towards}, plays a vital role in identifying the underlying cause behind the occurrence of the incident. By conducting a root cause analysis, the on-call engineers can uncover the primary issues that caused the incident and take appropriate corrective actions to prevent similar incidents from recurring in the future. This task is crucial for effective incident resolution, enhancing system reliability, and improving overall incident response processes.

Although LLMs have demonstrated impressive performance in incident diagnosis tasks \cite{ahmed2023recommending} through model fine-tuning on the incident data, they still face several challenges when applied to the root cause analysis task. Firstly, the current fine-tuned model operates under the assumption that it can learn all the intricate details of past incidents. However, it is widely recognized that the large language model is susceptible to hallucination (producing distorted or exaggerated facts), as it cannot accurately recall the details from the training data. Secondly, fine-tuning large language model is associated with high costs and may even be infeasible for certain cutting-edge models with an extremely large number of parameters such as GPT-4. Lastly, the fine-tuned model struggles to address the issue of staleness, where emerging knowledge makes previous information obsolete. It is challenging to update the LLM with recent knowledge unless it is continuously fine-tuned with latest data. Consequently, this limitation hinders the model's capacity to seamlessly ingest new knowledge.

To address the aforementioned challenges, we propose an in-context learning based method for the root cause analysis task. Rather than relying on fine-tuning with incident management data to acquire domain-specific knowledge, we directly include relevant historical incidents as in-context examples to equip the LLM with knowledge from the incident management domain. 
When applying in-context learning to our task, several considerations and decisions need to be taken. Specifically, given the high cost associated with fine-tuning a LLM, is it possible to achieve comparable performance in the RCA task using a vanilla LLM model without fine-tuning? (RQ1) Is it feasible to employ a traditional retrieval augmented approach
to enhance the performance of a vanilla GPT model in the RCA task?  (RQ2) How does the in-context learning method help the vanilla LLM in root cause analysis task? (RQ3) Does having more in-context examples result in better performance? (RQ4) How does the performance vary with the relevance of the in-context examples (RQ5)? How does the ordering of in-context examples affect the performance (RQ6)? 

To answer these questions thoroughly, we conducted an extensive evaluation involving a large-scale dataset of 101,308 incidents across over a thousand services from \company{}, one of the largest cloud service providers. In addition to the commonly reported lexical and semantic evaluation metrics for such experiments, we also present the results from a human validation, where we sought the input of incident owners to assess the correctness and readability of suggested root causes. Since the original incident owners possess the highest level of expertise regarding their incidents, their evaluation provides valuable insights into the performance of the models. Our contribution can be summarized as:
\begin{enumerate*}[label=(\roman*)]
    \item This work represents a pioneering effort in showcasing the practicality of cutting-edge Large Language Models (LLMs) like GPT-4 for accomplishing root cause analysis tasks without the need for fine-tuning, achieved through an innovative in-context learning approach. (Section~\ref{sec:methodology})
    \item We present a rigorous and large-scale study in \company{} on over 100,000 incidents from 1000+ cloud services with multiple evaluation metrics. The proposed in-context learning approach outperforms the fine-tuned GPT-3 model by an average of 24.7\% across all metrics. (Section~\ref{sec:exp_results})
    \item Our human study with the actual incident owners of production incidents serves as compelling evidence for the effectiveness of the proposed approach, showcasing notable improvements of 43.5\% in correctness and 8.7\% in readability. (Section~\ref{sec:human_eval})
\end{enumerate*}
The key takeaways of our work is 
\begin{enumerate*}[label=(\roman*)]
    \item Our proposed in-context learning RCA approach not only circumvents the high cost of fine-tuning incident data but also achieves even better performance compared to the existing fine-tuned LLMs.
    \item In comparison to the traditional retrieval augmentation approach, in-context examples can serve not only as task exemplars but also facilitate the integration of  domain knowledge into vanilla LLMs, resulting in a substantial performance improvement.
\end{enumerate*}

To reproduce our proposed approach, we will make the source code publicy availabel at \url{http://to_be_released}. However, due to privacy concerns with customer data in our dataset, we cannot release the full dataset. Instead, we include a detailed guide in our code repo on how to create a similar dataset step-by-step, along with some sample data for reference. Researchers can then use this guidance and the sample data to build their own datasets and apply our code to replicate our results.


\section{Background} \label{sec:overview}
In this section, we begin with an introduction to the root cause analysis task and the recent developments in LLM models. Following that, we delve into a thorough discussion of the research questions and the human evaluation conducted in this study.
\subsection{Incident Root Cause Analysis}
In large-scale cloud services, it is inevitable to encounter production incidents that can significantly impact the customer experience and incur substantial engineering resources for troubleshooting. The incident life-cycle typically consists of four stages: detection, triaging, diagnosis and mitigation.
In the incident diagnosis stage, root cause analysis plays a critical role in identifying the underlying cause of the reported incident. This process is complex and demands a significant amount of manual effort, as well as domain knowledge about the services involved. Incidents can stem from various issues, such as code bugs, dependency failures and infrastructure problems. The abundance of possibilities makes it challenging for On-Call Engineers (OCEs) to pinpoint the exact cause of the incidents. Errors made during the root cause analysis not only result in additional effort and labor but also have a direct impact on customers and revenue. It is essential to avoid such human errors to minimize disruptions and provide a better experience to customers. Figure \ref{fig:example} illustrates a real incident from a service, displaying the customer-provided title and summary, along with the actual root cause.

\begin{figure}[t!]

\fbox{\begin{minipage}{0.98\columnwidth}\small
\textbf{Title:} Completion Mismatches Between Service-A and Service-B

\vspace{2pt}
\textbf{Summary:} Recent availability issues in Learner Service have resulted in lost data for Service-A completions for many users. This ticket will track those incidents.

\vspace{2pt}
\textbf{Reference root cause:}
Service-A sync job was not able to handle dependent service unavailability.
\end{minipage}}
\centering
\caption{A sample production incident.}
\label{fig:example}
\vspace{-0.15in}
\end{figure}

\subsection{The Promise of LLMs}

In recent years, Large Language Models (LLMs) like GPT-4 \cite{openai2023gpt4} have become a prominent trend in natural language processing. With billions of model parameters, LLM models are trained on meticulous collections of textual content, ranging from books to web texts and Wikipedia articles. This comprehensive learning process enables LLM models to comprehend a wide range of prompts and questions, resulting in higher accuracy and precision when handling complex tasks.
GPT models, particularly the GPT-4 model, have demonstrated their superiority over state-of-the-art models in various NLP tasks, including machine translation, question-answering, and close tasks. Furthermore, GPT-4 has shown that unsupervised language models trained with sufficient data can perform at the same level as fine-tuned models with only a few examples of new tasks, leading to significantly improved performance. 
With enhanced capabilities, GPT-4 allows a token limit of 32,000 (equivalent to 25,000 words), a substantial increase compared to GPT-3.5's previous limit of 4096 tokens. This enhancement empowers GPT-4 to handle even complex questions by synthesizing information from diverse sources. 

\subsection{Research Questions}

We investigated several OpenAI GPT-3.x and GPT-4 models to generate root causes for the incidents without model fine-tuning. This leads to several RQs.

\smallskip
\noindent{\underline{\em RQ1} }\emph{Given the high cost associated with fine-tuning a LLM, is it possible to achieve comparable performance in the RCA task using a vanilla LLM model without fine-tuning?}

\noindent Vanilla GPT models, which lack training with incident management data and domain knowledge, are not expected to perform well in zero-shot settings. On the other hand, even though the fine-tuned model can acquire incident domain knowledge from the training data, it still faces the burden of high training and maintenance costs. To tackle these issues, we explore the in-context learning approach, which integrates LLMs with in-context examples as task exemplars and augmented domain knowledge, eliminating the need for time-consuming fine-tuning. To demonstrate the effectiveness of the in-context learning approach, we compare its performance with the fine-tuned model in the root cause analysis task.

\smallskip
\noindent{\underline{\em RQ2} }\emph{Is it feasible to employ a traditional retrieval augmented approach to enhance the performance of a vanilla GPT model in the RCA task?}

\noindent Retrieval-augmented approaches \cite{lewis2020retrieval, liu2022knowledge, guu2020retrieval} have emerged as a powerful technique to enhance the performance of LLMs by incorporating external documents. This integration allows language models to leverage external domain knowledge, leading to improved contextual understanding. However, some approaches \cite{borgeaud2022improving, wu2022memorizing}, require fine-tuning a specific decoder to leverage the retrieval-augmented knowledge, which contradicts our motivation to avoid fine-tuning the model. On the contrary, there are other methods that directly integrate the documents into the model input, wherein the augmented document can only furnish domain knowledge but lacks the functionality of a task exemplar. Furthermore, using chunked retrieval documents may reduce the effectiveness of the retrieval compared to the format of in-context examples. It also remains a question whether chunked retrieval documents can surpass the performance of the format of in-context examples. To assess the significance of in-context examples, we conducted a comparison between our in-context learning approach and the traditional retrieval augmentation method. In the latter, we divided incident details, comprising incident title, summary and root cause, into chunks, disregarding the original format of incident fields. The resulting document was presented as a sequence of chunks rather than as in-context examples. This comparison allowed us to illustrate the importance of using in-context example format while we maintain content consistency for the two approaches.

\smallskip
\noindent{\underline{\em RQ3} }\emph{How does the in-context learning method help the vanilla LLM in root cause analysis task?}


\noindent In-context learning methods have proven their capability to bridge the domain knowledge gap for LLMs by utilizing demonstrated examples in many domains. In this research, we aim to explore how this method can aid the vanilla LLM model (without fine-tuning) in enhancing the root cause analysis task. We utilize four different GPT models with varying capacities and observe their performance with in-context examples to compare with their zero-shot version for the root cause analysis task. 

\smallskip
\noindent{\underline{\em RQ4} }\emph{Does having more in-context examples result in better performance?}

\noindent When considering the significance of in-context examples for LLM, especially without fine-tuning on domain-specific data, the question arises of whether more in-context examples lead to better results. Ideally, we expect the LLM to be capable of analyzing the integrated in-context examples and distinguishing the useful ones. Our aim is to find a balance between the quantity and quality of these examples. To achieve this, we conduct experiments using various sizes of in-context examples and observe how it affects the root cause analysis task. Moreover, LLMs like the GPT-4 model have been developed to accommodate an exceptionally large number of input tokens. For example, the GPT-4-32K model can handle up to 32 thousand tokens in its prompt. This substantial increase in capacity significantly enhances the LLMs' ability to incorporate more background information, thereby further improving their contextual understanding for specific tasks. In our case, we utilized the GPT-4 model in our experiment, testing both the 8K and 32K prompt limits. This allowed us to integrate a greater amount of historical incidents as reference data, serving as background knowledge for the LLM.

\smallskip
\noindent{\underline{\em RQ5} }\emph{How does the performance differ between highly relevant in-context examples and irrelevant (random) examples?}

\noindent In-context examples typically serve two main functions. Firstly, they can be utilized as task demonstrations. By providing input and output through in-context examples, LLMs can learn how to perform the task based on these examples. Secondly, the content of in-context examples can also provide LLMs with domain-specific knowledge to tackle new incidents. In this paper, we aimed to investigate whether the relevance between in-context examples and new incidents plays a crucial role in determining the performance. We compare the in-context examples most relevant to the new incident with randomly selected examples.

\smallskip
\noindent{\underline{\em RQ6} }\emph{How does the ordering of in-context examples affect the performance?}

\noindent Previous research has revealed that LLMs are influenced by the arrangement of in-context examples. The main question we aim to address is whether we should place our in-context examples closest to the task description (at the beginning of prompt) or the new incident (bottom of prompt). To examine the impact of the order of in-context examples, we compare the performance of three different settings. Firstly, we present the examples in descending order of relevance, with the most relevant example coming first and the least relevant example last, relative to the new incident. In the second setting, we arrange the examples in ascending order, meaning that the most relevant example is positioned closest to the new incident. Lastly, we select the top k most relevant examples and then randomize their positions to study the effect of this arrangement.

\section{Methodology} \label{sec:methodology}
We present our in-context RCA approach that uses the in-context examples to enhance the performance of LLM. First, we provide an overview of our approach in Section \ref{sec:overall_arch}. Then we delve into the details of the data preparation and in-context example extraction in Section \ref{sec:data_prep} and Section \ref{sec:ice_extract}. Last, the root cause generation step is described in Section \ref{sec:rc_gen}.

\subsection{Overall Architecture} \label{sec:overall_arch}

\begin{figure}[t]
    \centering
    \scalebox{0.99}{
    \includegraphics[width=0.49\textwidth, trim={2.0cm 1.6cm 0.1cm 0.0cm}]
    {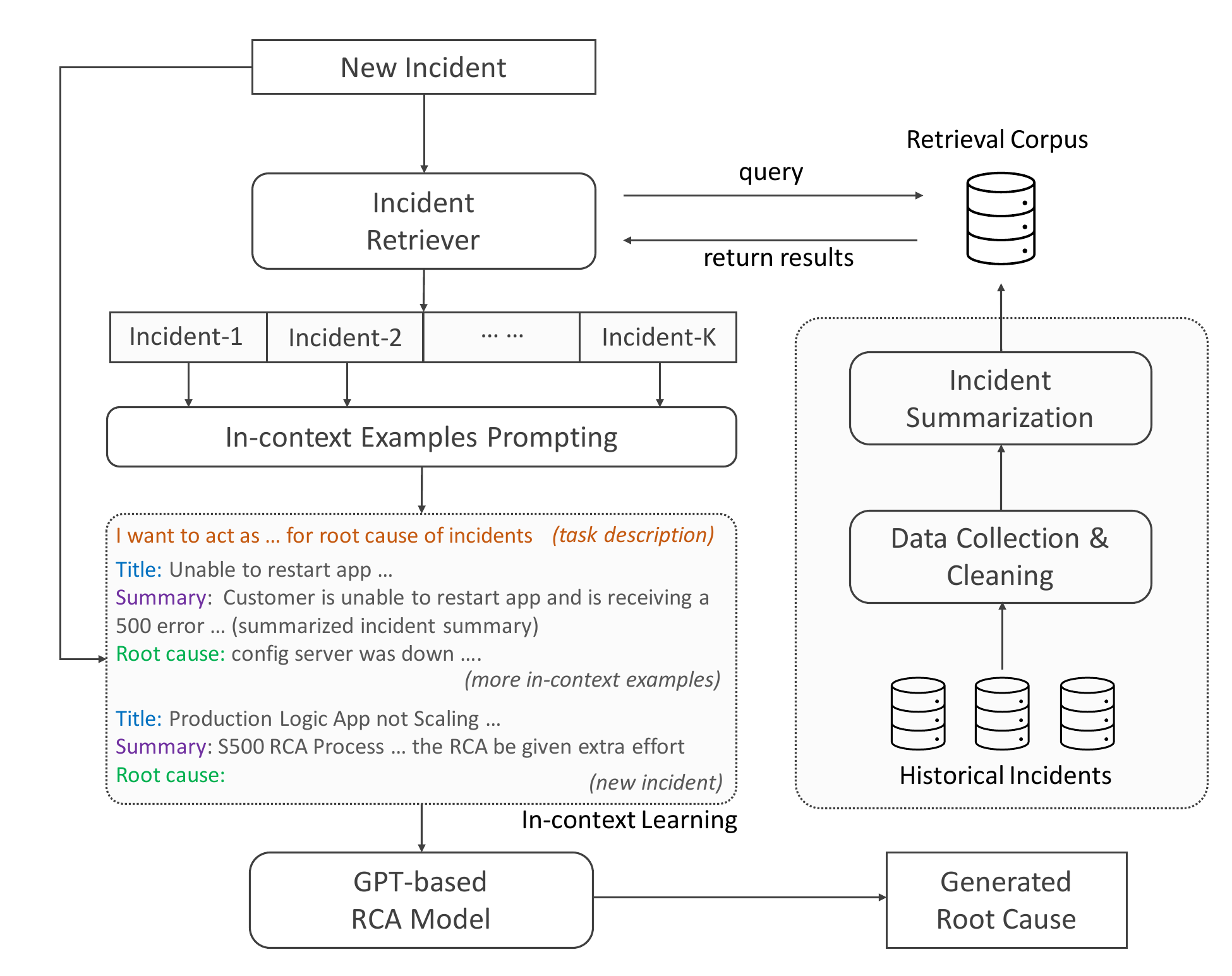}}
    \caption{Overview of our In-context Learning RCA Framework}
    \label{fig:overview}
\end{figure}

Figure \ref{fig:overview} illustrates the overall architecture of our proposed approach. Initially, we gather incident data created between January 1, 2021, and September 30, 2022, from our incident database. The data is then cleaned by removing lengthy stack traces and embedded images. To avoid overwhelming amounts of incident details, we utilize GPT-35-turbo\footnote{https://platform.openai.com/docs/models/gpt-3-5} to summarize the incident summary and root cause for constructing the retrieval corpus and in-context examples. After summarization, we employ a sentence transformer model \cite{reimers-2019-sentence-bert} to generate embedding vectors for each incident's summarized summary. Subsequently, we construct a retrieval index using the Faiss library \cite{johnson2019billion}, enabling efficient similarity search based on these embeddings. When a new incident arises, we use its description as a query to find relevant incidents based on the retrieval index. The extracted incidents are then integrated into the prompt of the Large Language Model (LLM) in the form of in-context examples. Finally, we utilize the LLM, such as GPT-4, to generate the root cause based on the new incident description and all the provided in-context examples.

\subsection{Data Preparation} \label{sec:data_prep}
Our data preparation process involves two steps. Initially, we extract the data from the incident database using specific criteria. Next, we proceed to clean the incident samples.

\subsubsection{Data Collection}
Numerous incidents from different services and severities are detected and created at \company through both automated systems and human monitoring. A team of dedicated on-call engineers (OCEs) is always ready to address these incidents promptly, ensuring seamless service for our valued customers. To effectively manage this high volume of incidents, \company has developed a specialized platform tailored for reporting and handling such occurrences. This platform includes a comprehensive database that tracks all activities related to incident reporting, from data insertion, modification, to deletion, starting from the moment an incident is created until it is successfully resolved.
      
We gathered 101,308 incident data\footnote{Unfortunately, we are unable to share the incident dataset as it contains confidential information about the cloud services and infrastructure.}  created between January 1, 2021, and September 30, 2022 that had both a non-empty summary and root cause field. These incidents were part of the "Resolved" or "Mitigated" incidents, with severity levels ranging from 0 to 4, where level 0 represented the most severe incidents. Furthermore, we applied filters to include only incidents whose titles contained specific keywords like "ignore," "test," or "dummy." Once the filtering process was completed, we sorted the remaining data based on the creation date. We then selected the first 98,308 incidents for the retrieval set, 2,000 for the validation set, and 1,000 for the testing set. To enable a comparison with the fine-tuned model, we further refined the retrieval set by choosing the most recent 20,000 incidents to be used as the training set for fine-tuned baseline models.

\subsubsection{Data Cleaning}
Because of the urgency in resolving incidents, the OCEs typically don't adhere to a strict template when providing incident summaries and root causes. This leads to the content of these incidents being challenging to parse and understand using rule-based models. Moreover, both the summary and root cause often contain information presented in various formats, including screenshots, tables, stack traces and code snippets. However, since GPT-3.x or GPT-4\footnote{The GPT-4 model does have support for image input, but it is currently not available for public usage.} only support textual data, we had to exclude the images from the incident summaries and root causes. Therefore, we conduct the following steps for incident data cleaning:
\begin{enumerate*}[label=(\roman*)]
    \item eliminate lengthy stack traces. Based on our observations, we have encountered stack traces that exceed 10,000 tokens. These lengthy stack traces obviously cannot be accommodated within the LLM's prompt and tend to distract the LLM with unnecessary details. Our proposed solution involves using a regular expression pattern to search for function calls within the incident summary. This pattern allows us to identify lines that are part of the stack trace, specifically by recognizing all lines in the incident summary that contain at least one function call.
    \item remove Base64 image. The incident contains images encoded in Base64 format, which is an encoding algorithm used to convert images into readable strings. This allows the images to be saved or transmitted over a network without any data loss. However, as LLM cannot comprehend Base64 images, we address this issue by utilizing the BeautifulSoup library to parse the \textit{<img>} tag and subsequently remove these images from the incident.
\end{enumerate*}

\subsection{In-context Example Extraction} \label{sec:ice_extract}
In this stage, we construct the retrieval corpus to find relevant incident examples that share similarities with the new incident. The process involves two key steps. First, we use incident summarization to condense the length of the original incident, making it more manageable. Next, we create a retrieval index that facilitates efficient semantic search based on the incident descriptions. This allows us to effectively find and retrieve incidents that closely match the semantics of the new incident.

\subsubsection{Incident Summarization}


Before we delve into the approach of summarizing the incident summary and root cause, let's first explain why such summarization is necessary. Incidents often contain an overwhelming amount of details that may not be helpful in reflecting the core information of the incident. For instance, Figure \ref{fig:summ_example} illustrates how incidents can include detailed logs, which, if included as is, would occupy most of the space in the in-context examples and limit the number of examples that can be used. In extreme cases, if the length of the summary exceeds 8K tokens, not even a single example can fit into the prompt for GPT-4 model. Furthermore, incorporating excessive unnecessary information can diminish the retrieval module's effectiveness. Existing retrieval models are typically pre-trained on smaller queries, which poses significant challenges when searching through lengthy documents with extended queries. 


Based on our observations, we employ the prompt shown in Figure \ref{fig:summ_prompt} to summarize the incident summary and root cause using the GPT-3.5-turbo model\footnote{\url{https://platform.openai.com/docs/models/gpt-3-5}}. The objective of this prompt is to make the GPT model function as a software engineer and incorporate specific information into the summarized version, including: incident symptoms, references to external services, distinguishing features such as error codes, and contextual details like the service name. Additionally, we ensure that the length of the summarized version does not exceed 5-6 sentences and is written in the third person's tone. To apply the prompt to new incidents, we substitute the placeholders "${\textrm{description}}$" with the incident summary and root cause, respectively. An illustrated example of a summarized incident can be seen in Figure \ref{fig:summ_example}. In this example, the summarized incident includes the key symptom "ABC\footnote{To ensure the confidentiality of sensitive information, we employ the placeholders "ABC" instead of the actual service name and "XXX" for error codes and IDs.} job is stuck during execution and not creating new records," along with the associated error code and initial troubleshooting steps provided by customers. The summarization process helps reduce the original summary from 864 tokens to just 89 tokens, which can be highly beneficial in saving space for the LLM prompt.

\begin{figure}[t!]

\fbox{\begin{minipage}{0.98\columnwidth}\small
\textbf{Original Incident Summary (864 tokens):} 

Customer XXX/Severity A ICM Ticket Template: Project ID:XXX - PREMIER SUPPORT CASE Project Name:XXX Environment Name:XXX Application/Platform Versions:XXX (10.0.XXX.XXX) : PU 36 (X.X.XXXX.XXXXX) Case Info:XXXXXX - Open Unify Environment Urgency Reason: Job is critical to the customer, issue affects prod Potential Downtime window: NA Issue Description: <core description>

... ... ...


- Export job failed due to error(s) in ABCStagingWriter.execute(): ...

\vspace{2pt}
\textbf{Summarized Version (89 tokens):}

This customer is experiencing an issue with a ABC job that is stuck during execution and not creating new records. The customer has attempted to delete the job and recreate it, but is still having the same issue. The urgent reason is that the job is critical to the customer and affects production. The activity ID is ExcecutionID: XXX and the engineer has requested assistance in getting the job to complete. Kusto URLs have been captured for this incident.

\end{minipage}}
\centering

\caption{Example of Original and Summarized Incident}
\label{fig:summ_example}
\vspace{-0.15in}
\end{figure}

\subsubsection{Retrieval Index Building}
Once the incident has been summarized in the previous step, we proceed with encoding the incident summary using the Sentence Transformer (ST) model \cite{karpukhin2020dense}, specifically all-mpnet-bse-v2\footnote{https://huggingface.co/sentence-transformers/all-mpnet-base-v2}. This model has been fine-tuned on 32 datasets comprising 1 billion sentence pairs with a contrastive learning objective \cite{chuang2020debiased}. The ST model utilizes a sentence transformer encoder to convert sentences or paragraphs into a 768-dimensional dense vector space, which can help to encode the incident description into vector space. In our approach, we utilize this model to encode the concatenation of the title and summarized incident summary for historical incidents in the retrieval corpus. Likewise, for new incidents, we also use the same setting to generate the query embedding.

After obtaining the embedding vectors from the sentence transformer model, we utilize the FAISS \cite{johnson2019billion} library for efficient similarity search and clustering of the dense vectors derived from historical incidents. Since the incidents are represented as vectors, we can compare them using L2 (Euclidean) distances. The vectors with the lowest L2 distance from the query vector are considered similar to the query vector. The FAISS library employs a compressed representation of the vectors, eliminating the need to store the original vectors. Although this may lead to a slightly less precise search, the benefit lies in its ability to scale to billions of vectors in main memory on a single server. This scalability allows for efficient searches on our thousands of historical incidents, making it a feasible and practical approach for our requirements. The index generated by FAISS can be stored in the disk and loaded into memory for efficient search. 

\subsubsection{In-Context Examples Retrieval}

In this stage, our goal is to search for relevant in-context examples using the retrieval index that was generated in the previous step. Given an incident description $d$, the main objective of the retriever is to select the top-k most relevant incidents $\mathcal{D}_r=\{d_1, \dots, d_k\}$ from a large retrieval corpus $\mathcal{D}$, where $\mathcal{D}_r \subseteq \mathcal{D}$. To achieve this, we adopt the approach used in prior research \cite{lewis2020retrieval}. We utilize the sentence transformer model to encode the concatenation of the new incident title and summary, which produces an embedding representation known as the incident query vector. Afterward, we make use of the FAISS library to retrieve the top-k most relevant incidents based on the retrieval index created in the previous step. Once we have retrieved the relevant incidents, we combine their title, summary, and root cause to serve them as in-context examples.

\begin{figure}[t!]

\fbox{\begin{minipage}{0.98\columnwidth}\small
\textbf{Incident Summary Prompt:} 

I want you to act as an expert software engineer. Consider the following incident report was submitted on the IcM portal. 

Incident Description: \{description\}

Your task is to summarize this incident report. Focus on the following aspects of the incident:

$\cdot$ The symptoms of the incident that lead to this incident report 

$\cdot$ References to external services or tools that contain relevant information.

$\cdot$ Distinguishing features of the incident such as precise error codes, specifics from logs etc.

$\cdot$ Context of the incident such as the name of the service, region, etc.

Your summary should be at most 5-6 sentences and should be in third person. You must end your summary with <|endoftext|>.

Concise Summary:

\vspace{2pt}
\textbf{Incident Root Cause Prompt Summary:}

I want you to act as an expert software engineer.
Your task is to summarize the following root cause of an incident report. Your summary must clearly state what the root cause of the incident was.

Incident Root Cause:
\{description\}

Concise Summary:

\end{minipage}}
\centering

\caption{Summarization Prompt for Incident Summary and Root Cause}
\label{fig:summ_prompt}
\vspace{-0.15in}
\end{figure}
\subsection{Root Cause Generation} \label{sec:rc_gen}


To generate the root cause, our initial step involves constructing the prompt for the LLM using the retrieved in-context examples. As illustrated in Figure \ref{fig:incontext_prompt}, the prompt consists of the task definition, in-context examples and the description of the new incident. Initially, we define the root cause analysis task and prompt the LLM to act as a software engineer. Next, we present the retrieved in-context examples, organized with their titles, summaries, and root causes, with double new lines used to separate multiple incidents. It is worth noting that we utilize the summarized incident summary and root cause in the example to prevent excessive prompt space occupation while retaining the core information of the incident as reference knowledge for addressing new incidents. Following the list of in-context examples, we add the description of the new incident, including its title and summary. Notably, the incident summary used here is the original version, allowing for the inclusion of more details about the new incident. Finally, we conclude the prompt with the phrase "Root Cause:" to prompt the LLM to generate the root cause. Once the prompt is prepared, we utilize the OpenAI API\footnote{https://azure.microsoft.com/en-us/products/ai-services/openai-service} to call upon GPT models for generating the root cause. In particular, we set the temperature to zero to ensure a more deterministic output from the model. Additionally, we configure the completion length to 200 for the root cause generation.


\begin{figure}[t!]

\fbox{\begin{minipage}{0.98\columnwidth}\small
I want you to act as a software engineer to figure out the root cause of incidents. I will provide some examples to start with.\\\\
Title: \{incident title\}\\
Summary: \{summarization of incident summary\}\\
Root Cause: \{summarization of incident root cause\}\\
... ...\\
Title: \{incident title\}\\
Summary: \{incident summary\}\\
Root Cause: 
\end{minipage}}
\centering

\caption{In-context Examples Prompting}
\label{fig:incontext_prompt}
\vspace{-0.15in}
\end{figure}

\section{Experiment} \label{sec:experiment}

\subsection{Experiment Setup}

\subsubsection{Datasets and Labels}
As outlined in Section \ref{sec:data_prep}, our dataset comprises incident data collected from various services and severities at \company, totaling 101,308 incidents. To construct our data set, we select the first 98,308 incidents for the retrieval set, 2,000 for the validation set, and the remaining 1,000 for the testing set. Regarding the GPT3 fine-tuning model, we designate the last 20,000 incidents from retrieval set as the training set for fine-tuning the model. As for the labels, we use the extracted root cause from each incident data sample. 

\subsubsection{Evaluation Metrics}
We choose two types of quantitative metrics for evaluating our model: lexcial and semantic metrics. For lexical metrics, we opt for four classic metrics. Firstly, we employ the Rouge metric (Recall Oriented Understudy for Gisting Evaluation) \cite{lin2004rouge} to compare a candidate document against a set of reference texts. Specifically, we choose ROUGE-L \cite{lin2004rouge}, which considers sentence-level structural similarity and identifies the longest co-occurring n-grams using Longest Common Subsequence (LCS) statistics \cite{hirschberg1977algorithms}. Also, we utilize ROUGE-1 \cite{lin2004rouge} to consider the 1-gram matching . Moreover, we include METEOR (Metric for Evaluation of Translation with Explicit Ordering) \cite{banerjee2005meteor}, which is based on the harmonic mean of unigram precision and recall, and it also incorporates additional features like stemming and synonymy matching to enhance its accuracy. The last lexical metric we have chosen is GLEU \cite{wu2016google}, a deriative of BLEU (Bilingual Evaluation Understudy) \cite{lin2004orange}. The metric was proposed to overcome some undesirable properties when the BLEU metric is used for single sentences.

To assess our results based on the semantic meanings of words, we choose to use two semantic metrics instead of lexical metrics, as the latter only consider exact word matches without taking word meaning into consideration. The first semantic metric we utilize is BERTScore \cite{zhang2019bertscore}. It leverages pre-trained contextual embeddings from the BERT model \cite{devlin2018bert} to compare candidate and reference sentence words using cosine similarity. This method enables a more nuanced evaluation of semantic similarity. Next, we incorporate the NUBIA (NeUral Based Interchangeability Assessor) \cite{kane2020nubia}, a recently developed neural-based measure. NUBIA integrates various aspects of semantic evaluation, including semantic similarity, logical inference, and sentence legibility. It achieves this by exposing layers of pre-trained language models like RoBERTa STS \cite{liu2019roberta}, RoBERTa MNLI, and GPT-2 \cite{radford2019language}. This comprehensive approach allows us to obtain a more accurate and comprehensive evaluation of the semantic quality of our results.

\subsubsection{Baseline Methods}

Traditional RCA models usually depend on preselected features to predict the root cause label from a set list of predefined labels, which doesn't work for our goal of directly generating textual root causes. To address this, we've chosen to fine-tune three recent small or medium-sized language models for our task: OPT \cite{zhang2205opt}, CodeGen \cite{nijkamp2022codegen} and Bloom \cite{workshop2022bloom}. Specifically, OPT \cite{zhang2205opt} is decoder-only model that was pretrained using a diverse and extensive dataset compiled from publicly available sources on the internet, encompassing a wide range of topics and text types to ensure comprehensive language understanding and generation capabilities. CodeGen \cite{nijkamp2022codegen} is an autoregressive language model for program synthesis trained sequentially on the Pile, BigQuery, and BigPython datasets to gain better capability of program synthesis. BLOOM \cite{workshop2022bloom} is a decoder-only Transformer language model that was trained on the ROOTS corpus, a dataset comprising hundreds of sources in 46 natural and 13 programming languages. 

We also select the GPT-3 model as the sole fine-tuned large language model since it's the only GPT modelone for which we have the capacity to fine-tune using a large amount of incident data. However, even for the GPT-3 model, the fine-tuning process demands 16 V100-32GB GPUs, and during inference, 4 V100-32GB GPUs are required. Moreover, we choose four GPT models, namely GPT-35-turbo\footnote{\url{https://platform.openai.com/docs/models/gpt-3-5}}, Text-davinci-003, GPT-4 \cite{openai2023gpt4}, and GPT-4-32K \cite{openai2023gpt4}, as our baseline models. We compare their zero-shot model output with the results obtained from our in-context learning approach. Lastly, we adopt the traditional retrieval augmentation method \cite{lewis2020retrieval}, wherein we split the incident description into chunks and use these chunks as retrieval documents, which are then compared to our in-context examples. To enable these models to tackle root cause analysis, we fine-tuned them on our incident management dataset using the same approach as for the GPT-3 model. The key difference is that we trained them in the traditional way, predicting the next token for both the incident description and the root cause. However, with GPT-3, we could only fine-tune the root cause section due to its fine-tuning pipeline limitations.

\subsection{Experimental Results} \label{sec:exp_results}

\subsubsection{Given the high cost associated with fine-tuning a LLM, is it possible to achieve comparable performance in the RCA task using a vanilla LLM model without fine-tuning? (RQ1)}

Table \ref{table:utility} presents the effectiveness of our in-context learning approach using vanilla GPT models. We conducted a comparison based on four GPT backbone models against four fine-tuned models, including CodeGen, OPT, Bloom and GPT3, on a training set of 20,000 examples. Due to the high demand for GPU resources in fine-tuning GPT models, we could only fine-tune the models that smaller than GPT-3 model with our dataset. We employed an in-context learning model with a few-shot examples setting, allowing a maximum of 20 instances, as it demonstrated superior performance on our development dataset compared to other configurations. From the results, we can conclude the following:  1) 
Our GPT-4 model shows remarkable performance, outperforming CodeGen, OPT, and Bloom by 63.85\%, 18.15\%, and 18.21\% respectively, across six metrics on average. It also exceeds the GPT-3 fine-tuned model by 24.77\%. Notably, it achieves a 38.22\% performance improvement in ROUGE-L and 7.50\% in the Nubia metric, showcasing its superior performance compared to the fine-tuned LLM for root cause analysis. 2) The performance of GPT-35-turbo still falls short of the fine-tuned model, indicating that the in-context learning approach still relies on the potency of the LLM to fully leverage the in-context examples. 3) GPT-4-32K achieves results similar to the GPT-4 model, which can handle a maximum of 8K input tokens. This is primarily because we utilize 20 in-context examples, which do not exhaust the 8K input limit of the GPT-4 model. 4) The OPT and Bloom models notably exceeded the performance of the CodeGen model and even slightly surpassed GPT-3, with improvements of 5.39\% and 5.44\% respectively. This enhanced performance can be largely attributed to the distinct training methodologies we implemented.

We carried out significance testing for our proposed model as well. In analyzing the Rouge-L scores and comparing the GPT-4 model with baseline models, we were able to confidently reject the null hypothesis (H0) because the p-value is significantly lower than 0.05. For instance, the p-value when comparing our GPT-4 model to the fine-tuned GPT-3 model stands at 1.64e-11. Such a result strongly suggests that the observed differences in performance are not just random occurrences, but are due to fundamental differences in the models themselves.

\begin{table*}[t]
\centering
    \scalebox{0.9}{
    \begin{tabularx}{0.8\textwidth}{cc *{6}{Y}}
    \toprule
     &  ROUGE-L  &  ROUGE-1  &  METEOR  &  GLEU  &  BERTScore  &  Nubia  \\
     \midrule
     CodeGen           	                & 10.32 & 14.26 & 11.78 & 3.39 & 81.45 & 35.13 \\
    OPT           	                    & 15.75 & 19.12 & 16.33 & 6.14 & 83.09 & 42.65 \\
    Bloom           	                & 14.68 & 18.81 & 17.67 & \textbf{6.46} & 83.01 & 40.94 \\
    
    \midrule
    GPT3 Fine-tuned	                    & 14.39 & 17.18 & 16.16 & 5.83 & 82.27 & 40.91 \\
    \midrule
    GPT-35-turbo w/ ICL	                & 12.33 & 17.47 & 17.38 & 4.49 & 81.84 & 37.28 \\
    Text-Davinci-003 w/ ICL	            & 18.01 & 23.72 & 19.51 & 5.70 & 84.50 & 43.13 \\
    GPT-4 w/ ICL	                    & \textbf{19.89} & \textbf{26.08} & 22.40 & 6.37 & 84.91 & 43.98 \\
    GPT-4-32K w/ ICL	                & 19.86 & 26.05 & \textbf{22.41} & \textbf{6.39} & \textbf{84.96} & \textbf{44.19} \\
    \bottomrule
    \end{tabularx}}
		\caption{Comparison between fine-tuned model and in-context learning models w/o fine-tuning}
		\label{table:utility}
\end{table*}

\subsubsection{Is it feasible to employ a traditional retrieval augmented approach to enhance the performance of a vanilla GPT model in the RCA task? (RQ2)}
To demonstrate the superiority of in-context learning, we conducted a comparison with the traditional retrieval augmentation method, which involved chunking historical incident details. In this process, we combined incident details, including their title, summary, and root cause. Subsequently, we split these incidents into chunks, each containing 128 tokens, and constructed a retrieval index using the same sentence transformer model. For each new incident, we retrieved the top-k most relevant chunks from the retrieval corpus, and we experimented with four different chunk number settings, varying from 10 to 40 chunks. The performance of the chunked retrieval method was then compared to that of our in-context learning model, and the results are shown in Table \ref{table:retr_chunk}. Our findings revealed that our in-context learning model outperformed the chunked retrieval model. With 30 shots, our model achieved an average improvement of approximately 22.37\% across all six metrics\footnote{Due to space limitations, we only presented four metrics in Table \ref{table:retr_chunk}}. Additionally, we observed that the performance of the chunked retrieval model consistently increased until the chunk size reached 30, but it started to decline for chunk sizes larger than 30.

\begin{table}[t]
\centering
    \scalebox{0.9}{
    \begin{tabularx}{0.49\textwidth}{cc *{4}{Y}}
    \toprule
     &  ROUGE-L  &  METEOR &  GLEU  &  Nubia  \\
    \midrule
    Chunked (10 shots)	        & 13.87 & 17.92 & 4.90 & 38.48 \\
    Chunked (20 shots)	        & 14.01 & 17.31 & 4.77 & 39.88 \\
    Chunked (30 shots)	        & 14.22 & 17.50 & 4.93 & 39.73 \\
    Chunked (40 shots)	        & 14.01 & 17.11 & 4.73 & 40.18 \\
    \midrule
    In-context Examples	                    & \textbf{19.89} & \textbf{22.4} & \textbf{6.37} & \textbf{43.98} \\
    \bottomrule
    \end{tabularx}}
		\caption{Comparison between hunked incidents and in-context examples}
		\label{table:retr_chunk}
\end{table}
\begin{table*}[t]
\centering
    \scalebox{0.9}{
    \begin{tabularx}{0.8\textwidth}{cc *{6}{Y}}
    \toprule
     &  ROUGE-L  &  ROUGE-1  &  METEOR  &  GLEU  &  BERTScore  &  Nubia  \\
    \midrule
    GPT-35-turbo Zero-shot            & 8.25 & 13.64 & 14.03 & 3.21 & 80.60 & 33.78 \\
    GPT-35-turbo w/ ICL	                & 12.33 & 17.47 & 17.38 & 4.49 & 81.84 & 37.28 \\
    \hdashline
    \%gain for GPT-35-turbo	              & +49.45\% & +28.08\% & +23.88\% & +39.88\% & +1.54\% & +10.36\% \\
    \midrule
    Text-Davinci-003 Zero-shot	       & 11.08 & 16.77 & 14.39 & 3.62 & 82.63 & 37.81 \\
    Text-Davinci-003 w/ ICL	            & 18.01 & 23.72 & 19.51 & 5.70 & 84.50 & 43.13 \\
    \hdashline
    \%gain for Text-Davinci-003	                & +62.55\%	& +41.44\%	& +35.58\%	& +57.46\%	& +2.26\% & +14.07\% \\
    \midrule
    GPT-4 Zero-shot                     & 10.27 & 16.40 & 16.21 & 3.71 & 81.95 & 33.33 \\
    GPT-4 w/ ICL	                   & 19.89 & 26.08 & 22.40 & 6.37 & 84.91 & 43.98 \\
    \hdashline
    \%gain for GPT-4	                & +93.67\%	& +59.02\%	& +38.19\%	& +71.70\%	& +3.61\% & +31.95\% \\
    \midrule
    GPT-4-32K Zero-shot                     & 10.13 & 16.15 & 16.10 & 3.68 & 81.93 & 32.99 \\
    GPT-4-32K w/ ICL	                   & 19.86 & 26.05 & 22.41 & 6.39 & 84.96 & 44.19 \\
    \hdashline
    \%gain for GPT-4-32K	                & +96.05\%	& +61.30\%	& +39.19\%	& +73.64\%	& +3.70\% & +33.95\% \\
    \bottomrule
    \end{tabularx}}
		\caption{Comparison between zero-shot model and in-context learning model with 20-shot examples}
		\label{table:zero_shot}
\end{table*}
\begin{figure*}[ht]
\scalebox{0.97}{
\begin{subfigure}{.4\linewidth}
  \includegraphics[trim=1.5cm 0.1cm 0.1cm 0.1cm, width=\linewidth]{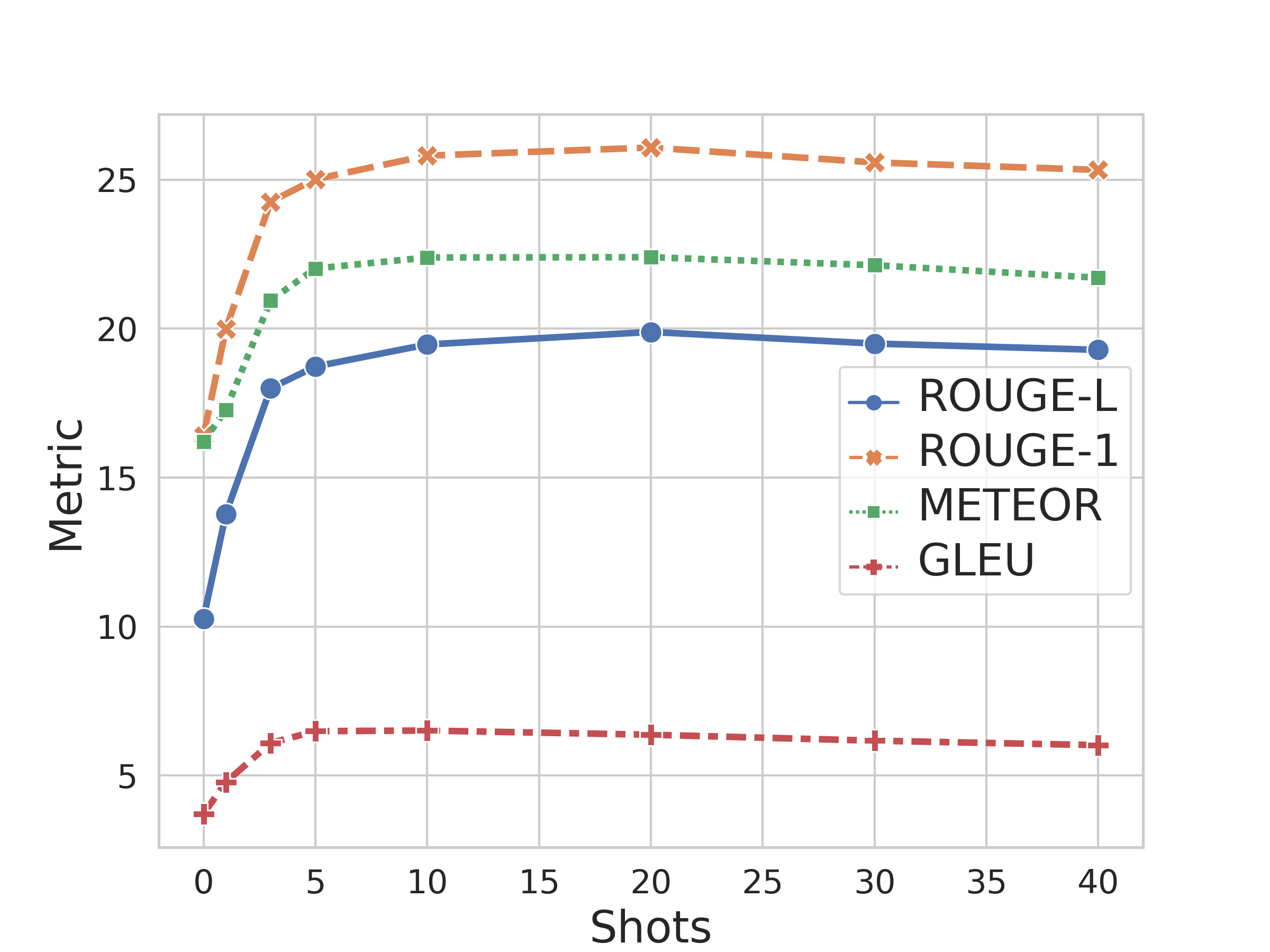}
  \caption{Lexical Metrics}
  \label{fig:fewshot_lexical}
\end{subfigure}

\begin{subfigure}{.4\linewidth}
  \includegraphics[trim=1.4cm 0.1cm 0.4cm 0.1cm, width=\linewidth]{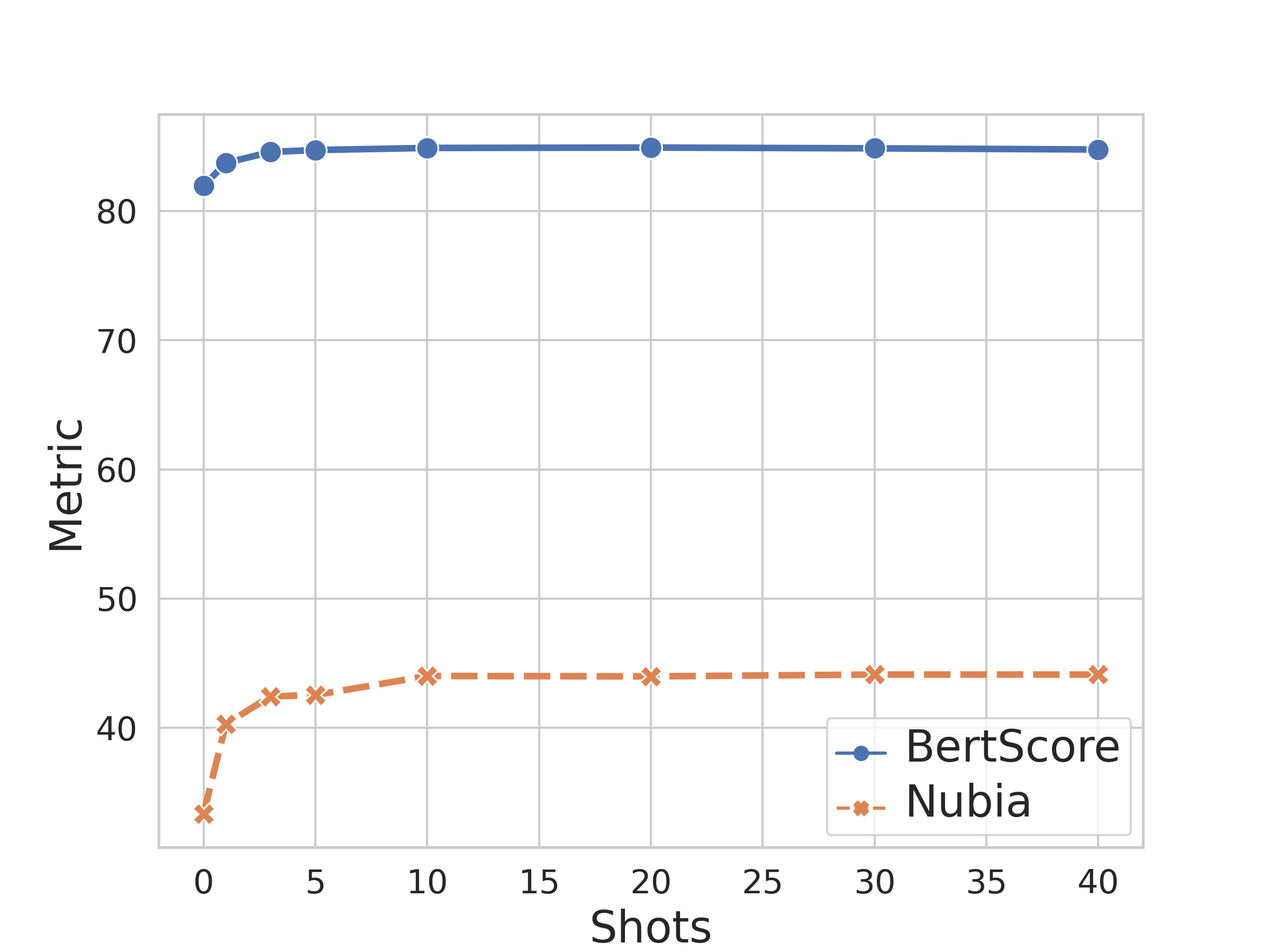}
  \caption{Semantic Metrics}
  \label{fig:fewshot_semantic}
\end{subfigure}
}
\caption{Performance for different few-shot examples}
\label{fig:few_shot}

\end{figure*}
\subsubsection{How does the in-context examples help the vanilla LLM in root cause analysis task? (RQ3)}

In Table \ref{table:zero_shot}, we conducted a comparison between the results of the zero-shot model and in-context learning models using 20 in-context examples, based on the four GPT baseline models. The findings reveal a substantial performance improvement with the in-context examples, showing significant gains of 49.69\% and 51.31\% for the GPT-4 and GPT-4-32K models, respectively. When comparing to the zero-shot model with the GPT3 fine-tuned model, we observe a performance drop of approximately 18.89\%. However, with 20 in-context examples, we achieve an improvement of 24.77\% over the fine-tuned model. Similar trends are seen for the GPT-35-turbo and Text-Davinci-003 models, with performance improvements of 25.53\% and 35.56\%, respectively, compared to the zero-shot model. Additionally, we noticed that the performance improvement for the two semantic metrics, BERTScore and Nubia, is notably smaller compared to the lexical metrics, particularly for the two GPT3.5 models.  Nonetheless, the GPT-4 models manage to achieve a noteworthy 30\% or higher performance gain in the Nubia metric compared to the GPT3.5 models, which show only 2-3 times less improvement. This underscores the superior ability of the GPT-4 models to enhance overall root cause analysis task through the utilization of in-context examples.

\subsubsection{Does having more in-context examples result in better performance? (RQ4)}
To address the research question, we initially employed the GPT-4 model on various few-shot settings, ranging from 0 to 40 shots. The results are presented in Figure \ref{fig:few_shot}, with lexical metrics displayed in Figure \ref{fig:fewshot_lexical}, and semantic metrics depicted in Figure \ref{fig:fewshot_semantic}. We discovered that both lexical and semantic metrics achieved optimal performance when the number of shots reached 20. Moving from 0-shot to 10 shots, we observed a significant improvement of 46.12\% on average across all the metrics. However, increasing the shots from 5 to 10 only resulted in a marginal 2.12\% improvement, and further increasing the shots to 20 showed an even smaller improvement of 0.18\%. Beyond 20 shots, we noticed some performance degradation, likely due to the inclusion of more irrelevant examples when using too many in-context examples.

Additionally, Table \ref{table:few_shot} presents a comparison between fixed 20-shot examples and full prompts that fill up to the token limit of the LLMs. It is evident that all the models exhibited worse performance when too many examples were included in the prompt. For instance, the GPT-4-32K model had an average of approximately 160 in-context examples, which proved to be sufficient to include both relevant and irrelevant examples. The presence of irrelevant examples had a detrimental impact, causing a performance decrease of around 10.2\%. One important observation is that the average number of in-context examples for the two GPT3.5 models is approximately 17.0, which is even lower than the 20-shots setting. However, some samples might still contain significantly larger in-context examples than the 20-shots setting, which could potentially impact the performance due to the presence of irrelevant examples.

\subsubsection{How does the performance differ between highly relevant in-context examples and irrelevant (random) examples? (RQ5)}

\begin{figure}[t]
\centering
\includegraphics[width=0.8\columnwidth]{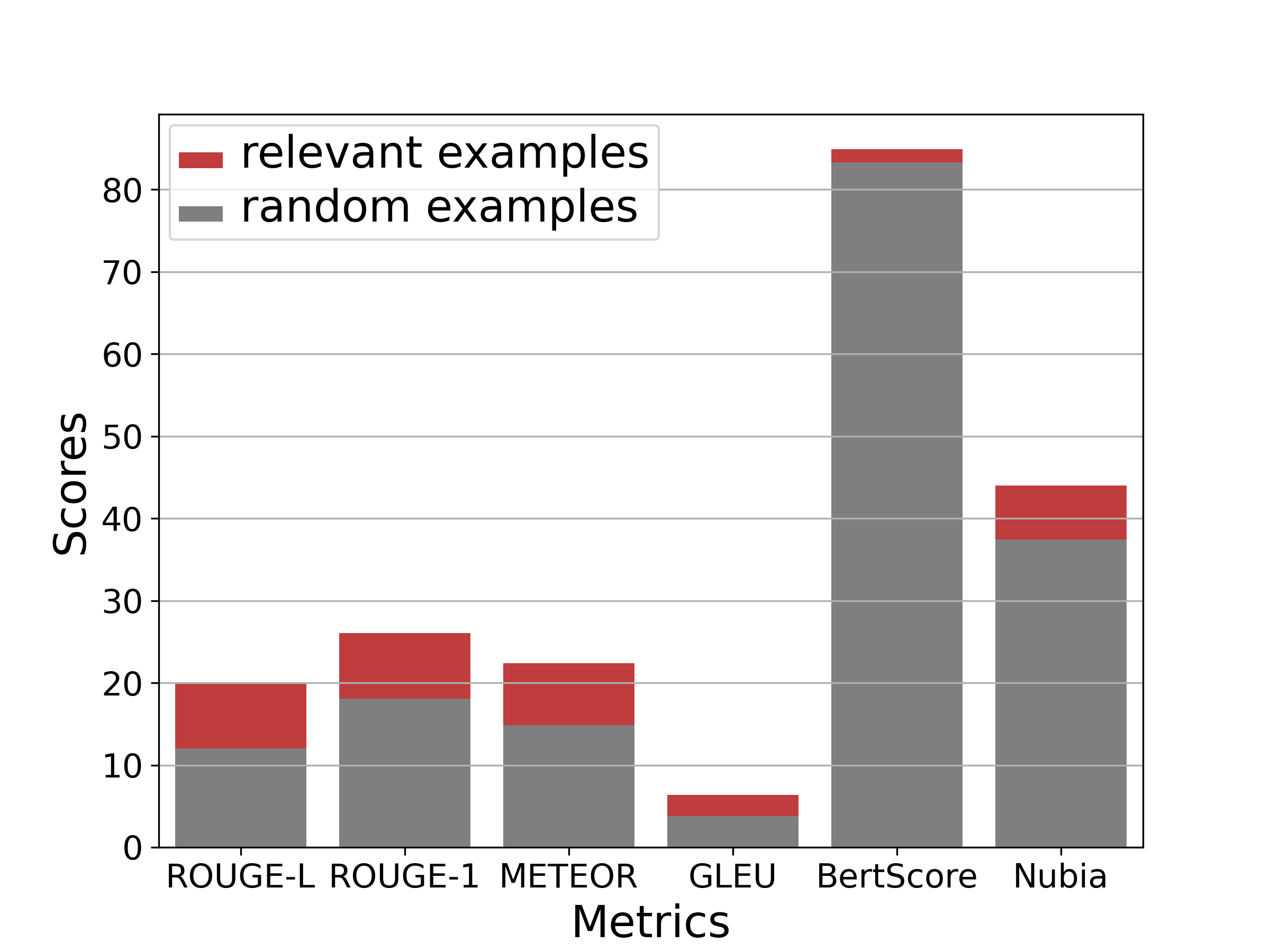}
\caption{Comparison between relevant and random in-context examples}
\label{fig:rand_shots}
\end{figure}

Figure \ref{fig:rand_shots} presents the comparison between the most relevant incidents and random incidents, both consisting of 20 in-context incident examples. It is evident that using the most relevant examples leads to a considerable performance improvement of approximately 41.2\% when compared to randomly selected incidents that lack any semantic relevance to the current incident. Notably, the ROUGE-L metric exhibits the most significant improvement, showing a remarkable 64.93\% boost, while BertScore, a metric with minimal variation between different methods, demonstrates a more modest 1.9\% difference. Additionally, Nubia shows an improvement of around 17.6\%, albeit the improvement for lexical metrics appears to be more significant than for semantic metrics. The reason for this observation can be attributed to two factors. Firstly, lexical metrics tend to have relatively lower values, which can amplify the percentage change. Secondly, lexical metrics heavily rely on word-level matching, favoring incidents that share more similar expressions, thereby providing a greater advantage in performance improvement. Moreover, when comparing the random examples to the zero-shot model in Table \ref{table:zero_shot}, we observe a 5.9\% performance improvement, which is significantly lower than the 49.7\% improvement achieved on relevant examples. These results indicate that the relevance of the in-context example contributes more than its functionality as a task exemplar in the RCA task.

\begin{table*}[t]
\centering
    \scalebox{0.9}{
    \begin{tabularx}{0.9\textwidth}{ccc *{6}{Y}}
    \toprule
    & &  ROUGE-L  &  ROUGE-1  &  METEOR  &  GLEU  &  BERTScore  &  Nubia  \\
    \midrule
    \multirow{2}{6.5em}{GPT-35-turbo}
    & 20 shots	                & 12.33 & 17.47 & 17.38 & 4.49 & 81.84 & 37.28 \\
    & full prompt ($\approx$ 17.0 shots)	            & 11.03 & 16.28 & 15.91 & 4.13 & 81.41 & 36.68     \\
    \midrule
    \multirow{2}{6.5em}{Text-Davinci-003}
    & 20 shots	            & 18.01 & 23.72 & 19.51 & 5.70 & 84.50 & 43.13 \\
    & full prompt ($\approx$ 17.0 shots)          & 17.73 & 23.49 & 19.53 & 5.45 & 84.44 & 42.41 \\
    \midrule
    \multirow{2}{6.5em}{GPT-4}
    & 20 shots	                   & 19.89 & 26.08 & 22.40 & 6.37 & 84.91 & 43.98 \\
    & full prompt ($\approx$ 37.6 shots)	                   & 17.18 & 23.13 & 19.90 & 5.41 & 84.37 & 41.43 \\
    \midrule
    \multirow{2}{6.5em}{GPT-4-32K}
    &20 shots	                   & 19.86 & 26.05 & 22.41 & 6.39 & 84.96 & 44.19 \\
    &full prompt ($\approx$ 160.0 shots)	                   & 17.13 & 23.09 & 19.66 & 5.29 & 84.36 & 41.48 \\
    \bottomrule
    \end{tabularx}}
\caption{Comparison between in-context examples with 20-shots and full prompt examples}
\label{table:few_shot}
\end{table*}

\begin{table*}[t]
\centering
\resizebox{.75\textwidth}{!}{%
\renewcommand{\arraystretch}{1.05}
\begin{tabular}{lcccccccccc}
\hline
\multicolumn{1}{c}{\multirow{2}{*}{Criteria}} & \multicolumn{2}{c}{GPT3 Fine-tuned} & \multicolumn{2}{c}{Text-Davinci-003} & \multicolumn{2}{c}{GPT-4} & \multicolumn{2}{c}{GPT-4-32K} \\
\multicolumn{1}{c}{} & Mean & Median & Mean & Median & Mean & Median & Mean & Median \\ \hline
Correctness & 1.72 & 1 & 1.69 (-1.9\%) & 1 & 2.47 (+43.5\%) & 2 & 2.13 (+23.6\%) & 2 \\
Readability & 4.34 & 5 & 4.47 (+3.0\%) & 5 & 4.72 (+8.7\%) & 5 & 4.59 (+5.8\%) & 5 \\
\hline
                          
\end{tabular}
}
\caption{Correctness and readability scores assigned by the incident owners}
\label{table:human_eval}
\end{table*}
\begin{table*}[t]
\centering
\small
\scalebox{0.9}{
\begin{tabular}{p{0.07\textwidth} p{0.9\textwidth}}
\toprule
Title & BreakFix | Networking | Device: Cable Reseat is blocked and needs assistance. \\ \midrule
Summary & Parent GDCO Task (Sev 3): xxx .... Blocking Description: Report Incorrect Ops Model for Servicing Vendor Dell for ...  Block Reason:  xxx Logs Attached? False   KB Article Number:  Did Server show POST?    Alias of Senior Tech \\ \midrule
Groud Truth & Data is inherited incorrectly due to a bug based on other factors from spares and sourcing or other attributes about the Rack itself \\ \midrule
Model Prediction & Data was inherited incorrectly due to a software bug that was based on other factors from spares and sourcing or other attributes about the Rack itself. \\ \midrule
In-context Examples & BreakFix | Networking | Cable Reseat is blocked and needs assistance. The incident report refers to a Sev 3 issue with the Parent xxx Task with a blocking task, related to the Operations Model being wrong for the xxx. The Block Reason is provided, but no xxx logs were attached. The incident report also lists the Device Name, Device Type, Rack, Slot and other details for reference.  The incident was caused by the \textbf{incorrect inheritance of data due to a bug}. This bug was influenced by other factors such as \textbf{spares, sourcing, or attributes of the Rack itself}.  \\ \midrule

\end{tabular}}
\caption{Case study of our in-context learning approach.}
\label{tab:case_study}
\end{table*}
\subsubsection{How does the ordering of in-context examples affect the performance? (RQ6)}
\begin{figure}[t]
\centering
\includegraphics[width=0.8\columnwidth]{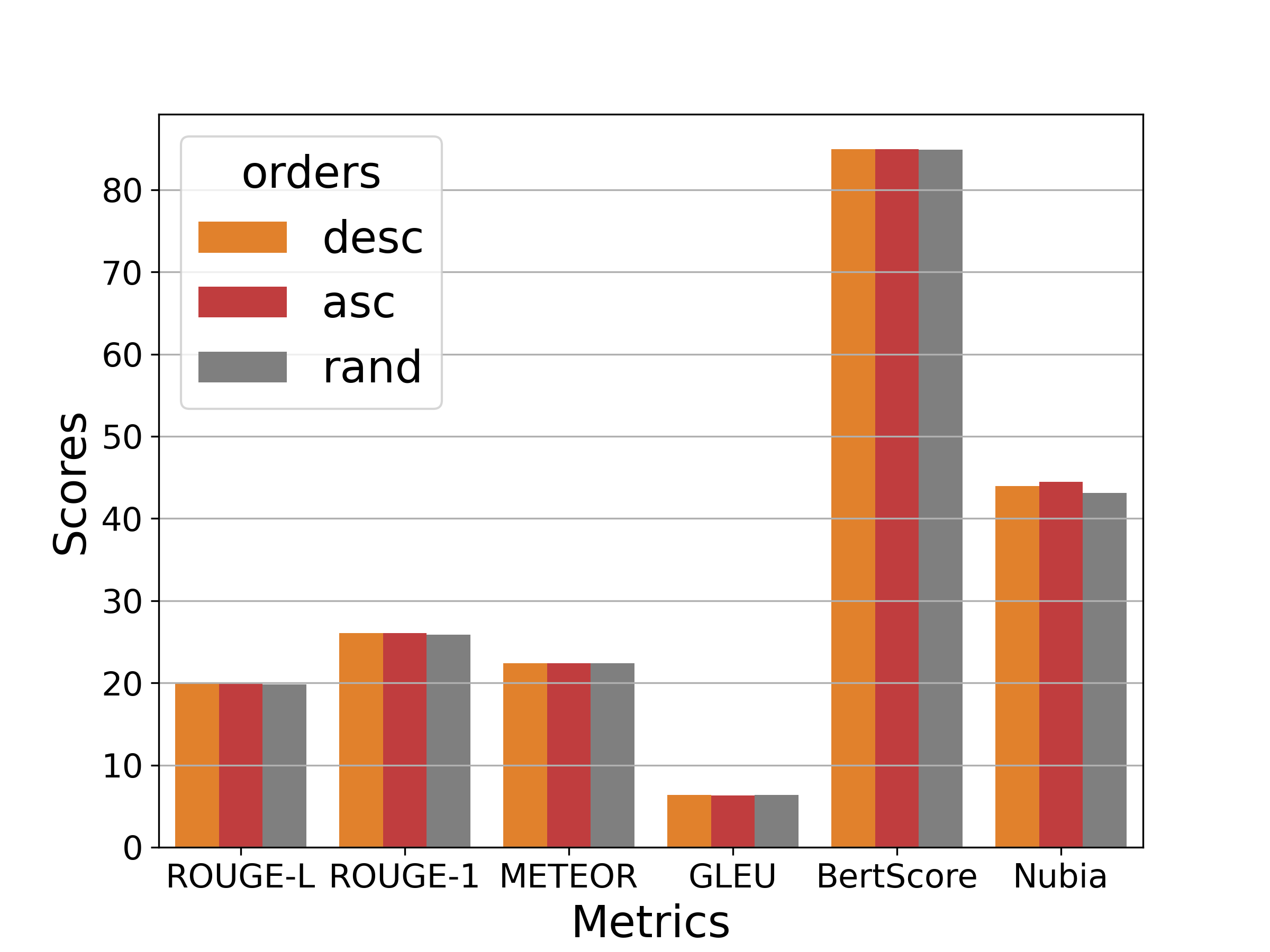}
\caption{Comparing different orders of the in-context examples}
\label{fig:order}
\end{figure}

Once the relevant incidents have been retrieved, a remaining research question is how to best arrange the order of these examples for achieving the best performance. To assess the impact of different ordering methods for the in-context examples, we conducted three experiments. Initially, we sorted the examples in descending order based on their relevance scores. Next, we tried the ascending order, and finally, we experimented with a randomized order. The results of these three ordering settings for all six metrics are depicted in Figure \ref{fig:order}, with a chosen in-context example size of 20. We observe that the order had minimal impact on performance. On average, the standard variance among the three settings was 0.12 across the six metrics. Notably, we observed a slightly larger variance for the Nubia metric, reaching up to 0.55, while all other metrics had a standard variance of less than 0.1.

\subsection{Human Evaluation} \label{sec:human_eval}
To showcase the human evaluation results, we conducted a random selection of 28 incidents from our incident pool. These incidents are spread across 11 different owning services, covering 8 different countries, and involving 18 distinct individual incident owners.
Table \ref{table:human_eval} showcases the human evaluation results, which were carried out by incident owners to ensure the accuracy of the assessments. The evaluations were based on two metrics: correctness and readability. The correctness metric aimed to determine whether the model offered helpful and relevant suggestions compared to the actual root-cause of the incidents. On the other hand, the readability metric assessed how easily readers could understand the generated text. The scoring system used ranged from 1 to 5, with 5 being the highest rating and 1 being the lowest. 
We utilized three in-context learning models (Text-davinci-003, GPT-4, and GPT-4-32K)  incorporating 20-shot examples for comparison with the fine-tuned GPT3 model. The results revealed that the GPT-4 model, enhanced with in-context examples, significantly outperformed the fine-tuned GPT3 model, scoring 43.5\% higher in terms of correctness. Moreover, the GPT-4 model exhibited an 8.7\% improvement in readability. The comparison also indicated that the Text-davinci-003 model slightly underperformed compared to the fine-tuned model by 1.9\%. This suggests that the in-context learning approach benefits from a more powerful model. Additionally, the use of only 20-shot examples hindered the GPT-4-32K model from leveraging its advantage of accommodating large prompt inputs, resulting in even poorer performance than the GPT-4 model.

We also discovered that the majority of the lowest-scoring incidents are connected to incidents that have a responsible incident linked to them. This linkage may lead to inaccuracies in the incident descriptions. Quoting a comment from the incident owner, ``I guess when dependency has issues, this shouldn't be treated as a normal case for the service in question". If we exclude such cases, we observe that the average score increases to 2.95, and nearly two-thirds (58.3\%) of the incidents achieve a score greater than 3. Only 12.5\% of incidents receive a score of 1. Moreover, crafting an entirely precise description of the root cause at the moment an incident is created is a non-trivial and challenging task. Our goal is to provide engineers with accurate guidance, so even a partially correct recommendation, such as a score of 3 out of 5, would be tremendously beneficial.

\subsection{Case Study} \label{sec:case_study}
Table \ref{tab:case_study} presents a case study showcasing the application of our proposed method using real-world incident data. The incident involves a problem referred to as the "Cable reseat is blocked" issue. The root cause of this problem was traced back to incorrect data inheritance, which occurred due to a bug related to spares and sourcing factors. Our model's prediction closely aligned with the ground truth, though the wording may vary slightly. By examining examples from our retrieval corpus, we identified similar incidents like the one displayed in Table \ref{tab:case_study}. The example shares the same issue but differ in their parent task and description. Nevertheless, our model effectively leveraged the root cause from these in-context examples to accurately predict the correct root cause for the given incident. This case serves as an illustration of the significance of using similar in-context examples. In contrast to fine-tuning the LLM, our approach relies entirely on extracted incidents, without concern for the generation of false information or reliance on outdated facts.
\section{Discussion} \label{sec:discussion}
\begin{figure}[t]
\centering
\scalebox{0.8}{
\includegraphics[trim=1.5cm 0.2cm 0.1cm 0.1cm, width=0.7\columnwidth]{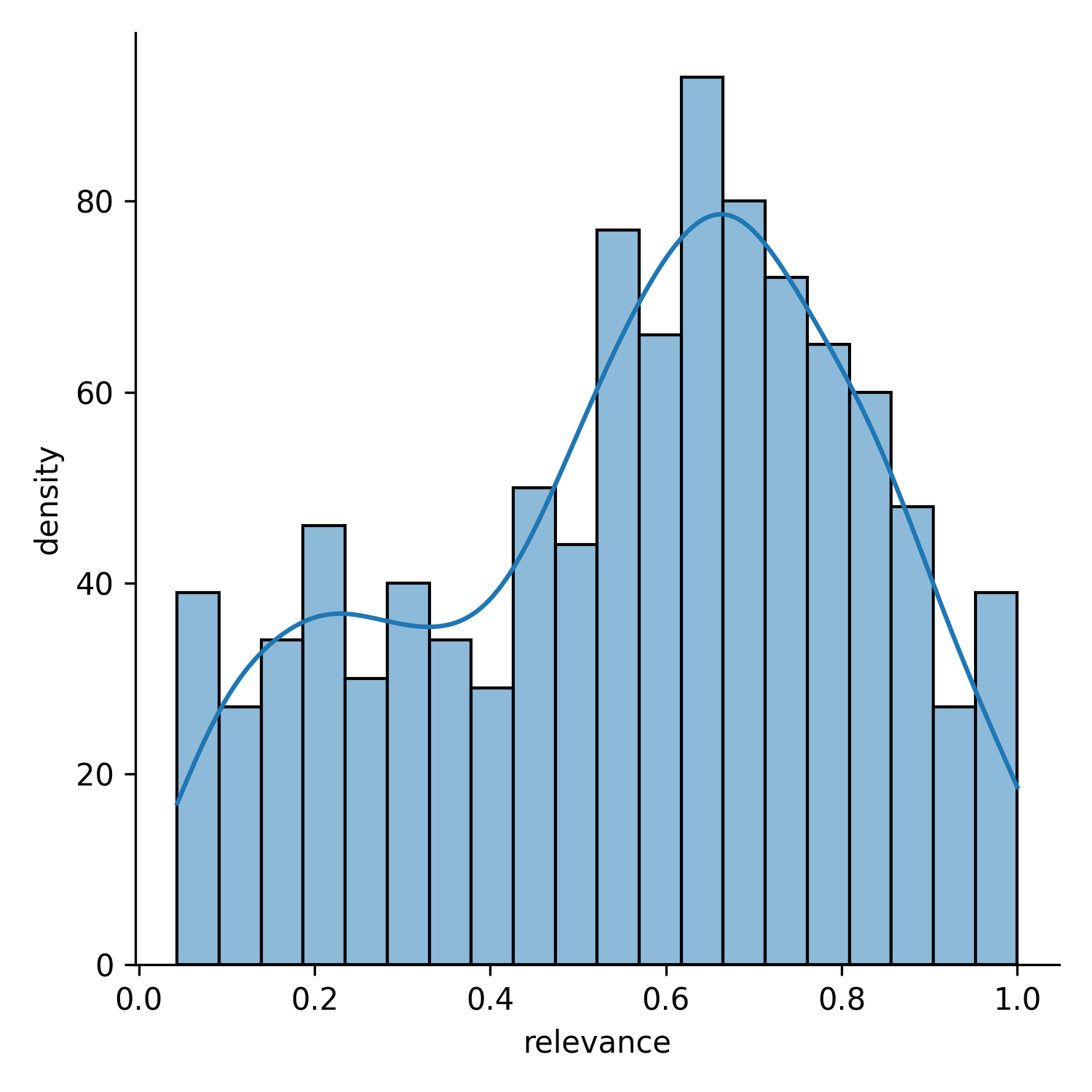}
}
\caption{Relevance distribution of historical incidents}
\label{fig:relevance}
\end{figure}

We address the limitations and future directions of our approach in the discussion section. 
Our approach has a limitation that its effectiveness heavily relies on the coherence of root causes among similar incidents. Consequently, we face challenges in dealing with completely new incidents where the same issue has not been previously resolved in other cases. This limitation arises from the fact that historical incident data cannot offer the necessary support for causal reasoning, especially when online diagnosis information is unavailable. However, practical systems often experience repeated faults for two primary reasons. Firstly, these faults can emerge from unavoidable issues like hardware malfunctions or unusual operational behaviors. For example, a connection timeout between two services might be caused by a disrupted network connection or a task in one service becoming unresponsive. Such issues can lead to problems reoccurring periodically. Secondly, the recurrence of these faults is often because implementing a fix in the product, or having customers apply a patch, usually takes a considerable amount of time. This delay can span several weeks to months, consequently extending the time it takes to fully resolve these faults. 
To uncover the ratio of  proportion of recurring versus novel incidents, we compared the relevance of current incidents against our historical incident corpus. Figure \ref{fig:relevance}  illustrates the relevance distribution based on relevance scores for the most relevant incident. We discovered that 18.8\% of incidents can find highly relevant incident from the history, with relevance scores exceeding 0.8. In contrast, 10.7\% of incidents were entirely unique, characterized by their highest relevance scores falling below 0.2. Another limitation is that our approach does not consider the age of the incident when referencing in-context examples. Therefore, when an incident's root cause becomes outdated, our current design lacks the capability to prioritize the new incident as the correct reference.

For future work, we can investigate how to incorporate online diagnosis tools into the approach and leverage LLMs to conduct step-by-step causal reasoning using the information obtained from these online tools. Furthermore, we can enhance the reasoning process by supplementing it with domain knowledge derived from the detailed diagnosis logs of historical incidents.
\section{Related Work} \label{sec:related_work}

\subsection{LLMs in Software Engineering}

In recent year, the emergence of LLMs has opened up new prospects in the software systems field, enabling various tasks such as program synthesis \cite{nijkamp2022codegen, jain2022jigsaw}, log analysis \cite{mastropaolo2022using}, vulnerability repair \cite{fu2022vulrepair}, software testing \cite{wang2023software}, and incident management \cite{ahmed2023recommending, chen2023empowering}. For example, Jain et al. \cite{jain2022jigsaw} propose an approach that enhances large language models with post-processing steps based on program analysis and synthesis techniques, resulting in improved performance of program synthesis. Mastropaolo et al. design LANCE system \cite{mastropaolo2022using} that utilizes fine-tuned T5 to automatically generate logging statements for Java methods. Similarly, VulRepair \cite{fu2022vulrepair} tool automatically suggests vulnerability fixes with a fine-tuned T5 model based on their vulnerability repair datasets. In constrast to previous studies, our approach harnesses the cutting-edge LLMs to generate root causes without requiring model fine-tuning, relying instead on the in-context learning method.

\subsection{Incident management}

Incident management within large cloud services has emerged as a popular research topic in the systems and software engineering communities. Several empirical studies have analyzed incidents and outages in production systems, specifically delving into incidents caused by particular types of issues \cite{leesatapornwongsa2016taxdc, alquraan2018analysis, gao2018empirical, zhang2021understanding} or issues arising from specific services and systems \cite{ghosh2022fight, liu2019bugs, yuan2014simple}.
Moreover, researchers have explored the use of machine learning and data-driven techniques to automate various aspects of the incident life-cycle, including triaging \cite{EmpiricalIcMICSE2019, ContinuousTriageASE2019, azad2022picking}, diagnosis \cite{nair2015learning, bansal2019decaf, luo2014correlating}, and mitigation \cite{jiang2020mitigate}. 
For root-cause analysis tasks, several research studies (e.g., TraceRCA~\cite{li2021tracerca}, CIRCA~\cite{li2022circa}, DiagFusion~\cite{zhang2023diag}, Eadro~\cite{lee2023eadro}) have been proposed for anomaly detection and root cause positioning. While these methods are useful to locate either categories of the root causes \cite{zhang2023diag,lee2023eadro} or to identify the potential problematic microservice \cite{li2021tracerca,li2022circa} to investigate, these does not provide a detailed description of the root cause. In contrast, we propose to generate the actual descriptive root cause information to guide On-Call Engineers (OCEs) into right direction by leveraging the power of LLMs. More specifically, our approach is designed as a generative task, which sets it apart from traditional RCA methodologies that treat problems as classification tasks. These traditional methods typically rely on predefined features to predict the root cause label from a fixed set of predefined root cause labels.

Recently, Ahmed et al. \cite{ahmed2023recommending} proposed a method to generate the textual root cause by fine-tuning GPT models using historical incident data. However, fine-tuning these models on state-of-the-art language models like GPT-4 poses significant challenges, such as requiring substantial GPU resources and incurring high maintenance costs for customizing the LLM for future use. Recently, Chen et al. \cite{chen2023empowering} developed a retrieval-augmented LLM model for root cause analysis but limited its application to specific service data, demanding domain-specific knowledge. In contrast, our approach leverages in-context learning with a substantial dataset comprising over 100,000 incidents. This allows us to support On-Call Engineers (OCEs) in resolving incidents without requiring fine-tuning or specific domain expertise in a broad context.

\section{Conclusion} \label{sec:conclusion}

In this paper, we present the effectiveness of utilizing cutting-edge language models like GPT-4 in root cause analysis task. We propose an in-context learning method that integrates historical incident knowledge into vanilla language models without the need for fine-tuning. Through extensive experiments on a large-scale incident dataset consisting of over 100,000 production incidents, we demonstrate that our in-context learning approach outperforms the fine-tuned GPT-3 model by an average of 24.8\% across six metrics. Additionally, the incorporation of in-context examples results in an impressive 49.7\% improvement over the zero-shot model. Human evaluation involving incident owners also indicates promising enhancements compared to the fine-tuned model, achieving a 43.5\% improvement in correctness. Considering the challenges of fine-tuning such massive incident data, our work provides valuable insights into utilizing cutting-edge language models effectively in our incident management domain without the necessity for fine-tuning.

\clearpage
\bibliographystyle{ACM-Reference-Format}
\bibliography{main}

\end{document}